\documentclass[conf]{new-aiaa}
\usepackage[utf8]{inputenc}

\usepackage{float}
\usepackage{gensymb}
\usepackage{graphicx}
\usepackage{amsmath}
\usepackage[version=4]{mhchem}
\usepackage{siunitx}
\usepackage{longtable,tabularx}
\title{Online Inertia Tensor Identification for Non-Cooperative Spacecraft via Augmented UKF}

\author{Batu Candan \footnote{PhD Candidate, Department of Aerospace Engineering, Iowa State University, AIAA Student Member 903147; dukynuke@iastate.edu}} \affil{Iowa State University, Ames, Iowa, USA, 50011.} 
\author{Simone Servadio \footnote{Assistant Professor, Department of Aerospace Engineering, Iowa State University, AIAA Member; servadio@iastate.edu}}
\affil{Iowa State University, Ames, Iowa, USA, 50011.}

\begin{document}

\maketitle

\begin{abstract}
Autonomous proximity operations, such as active debris removal and on-orbit servicing, require high-fidelity relative navigation solutions that remain robust in the presence of parametric uncertainty. Standard estimation frameworks typically assume that the target spacecraft’s mass properties are known a priori; however, for non-cooperative or tumbling targets, these parameters are often unknown or uncertain, leading to rapid divergence in model-based propagators. This paper presents an augmented Unscented Kalman Filter (UKF) framework designed to jointly estimate the relative 6-DOF pose and the full inertia tensor of a non-cooperative target spacecraft. The proposed architecture fuses visual measurements from monocular vision-based Convolutional Neural Networks (CNN) with depth information from LiDAR to constrain the coupled rigid-body dynamics. By augmenting the state vector to include the six independent elements of the inertia tensor, the filter dynamically recovers the target’s normalized mass distribution in real-time without requiring ground-based pre-calibration. To ensure numerical stability and physical consistency during the estimation of constant parameters, the filter employs an adaptive process noise formulation that prevents covariance collapse while allowing for the gradual convergence of the inertial parameters. Numerical validation is performed via Monte Carlo simulations, demonstrating that the proposed Augmented UKF enables the simultaneous convergence of kinematic states and inertial parameters, thereby facilitating accurate long-term trajectory prediction and robust guidance in non-cooperative deep-space environments.
\end{abstract}

\section{Introduction}
\lettrine{A}{utonomous} proximity operations for active debris removal and on-orbit servicing require accurate, high-rate relative navigation and prediction under significant modeling uncertainty. A central challenge is that non-cooperative targets are often tumbling and their mass properties are unknown or uncertain, causing model-based propagators to drift and eventually destabilizing tightly coupled guidance and estimation loops. While pre-flight inertia tensors are typically derived from CAD models, achieving high-fidelity representations that capture wiring harnesses, deployment configurations, and in-orbit mass redistribution is time-consuming and remains prone to mismatches; moreover, inertia can change over the mission lifetime due to fuel usage and configuration changes \cite{BOURABAH2023643, ceas2}.

A broad literature addresses on-orbit inertia identification using attitude information and excitation maneuvers. Many approaches leverage conservation of angular momentum and solve for inertia via constrained regression or least-squares formulations using reorientation sequences and onboard sensor data. More recent efforts emphasize that inertia identification is fundamentally limited by trajectory excitation and observability: estimation quality depends strongly on whether the collected motion sufficiently excites all axes, and practical methods must identify informative maneuver windows rather than relying on fixed tuning across an entire dataset \cite{CEAS1}. Kalman filter-based methods offer an attractive approach to online estimation because they naturally fuse dynamics and measurements sequentially. Beyond point-estimate formulations that assume Gaussian noise, set-membership and interval estimators have recently been proposed for inertia identification under bounded sensor uncertainty. In particular, zonotopic Kalman filtering has been used to propagate guaranteed inertia intervals using gyro and reaction-wheel sensing, and it is emphasized that maneuvers exciting rotation about all three axes are required for accurate and consistent interval recovery \cite{wangZKF2025}. Complementary to Bayesian filtering, data-driven system identification approaches estimate inertia from flight telemetry in closed-loop operation, for example using instrumental-variable estimators designed to remain consistent under milder noise assumptions \cite{nainerIFAC2018}. Finally, because many inertia estimators rely on angular acceleration, hybrid pipelines have been explored that denoise angular-rate measurements and obtain reliable angular accelerations via Savitzky--Golay differentiation before regression-based inertia estimation \cite{kimSGF2016}. These observability limitations can also imply that only inertia ratios are identifiable in some sensing configurations \cite{yoon2017}. In small-satellite and gyroless settings, rotational dynamics must be incorporated directly into the filter, and inertial parameters can be included as additional states; however, stable convergence requires careful covariance design and sufficient excitation \cite{gyroless}. These challenges become even more pronounced in relative navigation for non-cooperative rendezvous, where translational and rotational dynamics can be coupled through sensing geometry and where the target inertia is unknown. Vision-based relative navigation has enabled pose estimation of uncooperative objects, and several works have explored simultaneous pose and inertia recovery. Stereo-vision approaches can estimate pose and infer inertia ratios by augmenting inertial parameters and introducing pseudo-measurement constraints derived from the Euler equations, improving convergence but requiring access to angular acceleration information that is typically noisy or obtained via numerical differentiation \cite{PESCE2017236}. Other non-cooperative formulations estimate motion while treating inertia as an unknown parameter through interactive or multi-model filtering architectures \cite{YU2016479}. Recently, filtering frameworks with stronger convergence guarantees have been proposed, showing that with sufficiently rich rotational trajectories, relative pose and inertia can be recovered \cite{CREASER2024481, parre}.

Building on our prior work on vision-based marker/corner detection and relative pose estimation for uncooperative targets \cite{candan,candanMdpi}, as well as our earlier contributions on robust Kalman filtering and online covariance tuning \cite{candan_2022,batumtf,sokencandan}, this paper focuses on the remaining critical gap for long-horizon proximity operations: \emph{unknown target mass properties}. Specifically, we augment the existing CNN--LiDAR pose-estimation pipeline with an inertia-aware estimator by introducing an augmented UKF that jointly estimates the relative 6-DOF state and the six independent elements of the target inertia tensor. The key contribution is a practical, numerically stable parameter-estimation formulation that enables online recovery of the target inertia tensor without requiring pre-calibration, while retaining the same measurement interface used in our previous studies. The proposed approach is validated in a high-fidelity Blender-based ENVISAT simulation with realistic rendering, visibility labeling, and Monte Carlo trials, demonstrating simultaneous convergence of kinematic states and inertial parameters and improved long-horizon prediction under parametric uncertainty.

\section{Methodology}
\label{sec:methodology}
This section summarizes the proposed end-to-end pipeline for joint relative navigation and inertial-parameter identification. The overall architecture consists of (i) a high-fidelity translational and rotational dynamics model, (ii) a Blender-based synthetic dataset and camera model for ENVISAT, (iii) a CNN front-end that detects image-plane corner measurements, (iv) LiDAR-assisted depth association, and (v) an augmented UKF that estimates the relative 6-DOF pose together with the target inertia tensor and a depth-bias term. Extended derivations and implementation details are provided in \cite{candanMdpi, phdthesis}.

\subsection{Dynamics}
\label{subsec:dynamics}

\subsubsection{Absolute Chaser Motion}
The chaser is propagated under a two-body central gravity model. Denoting the chaser orbital radius by $\bar{r}$ and true anomaly by $\theta$, the planar equations of motion are
\begin{align}
\ddot{\bar{r}} &= \bar{r}\dot{\theta}^{2}-\frac{\mu}{\bar{r}^{2}}, \\
\ddot{\theta} &= -2\frac{\dot{\bar{r}}\dot{\theta}}{\bar{r}},
\end{align}
where $\mu$ is Earth's gravitational parameter. These states provide the reference angular rate and acceleration terms required for the relative equations in the local frame.

\subsubsection{Relative Translation in LVLH}
Relative translation is expressed in the chaser LVLH frame with unit axes $(\hat{\mathbf{i}},\hat{\mathbf{j}},\hat{\mathbf{k}})$. The relative position and velocity are
\begin{align}
\mathbf{r}_r &= x\hat{\mathbf{i}} + y\hat{\mathbf{j}} + z\hat{\mathbf{k}}, \\
\mathbf{v}_r &= \dot{x}\hat{\mathbf{i}} + \dot{y}\hat{\mathbf{j}} + \dot{z}\hat{\mathbf{k}}.
\end{align}
The nonlinear relative acceleration model is written as
\begin{align}
\ddot{x} &= 2\dot{\theta}\dot{y} + \ddot{\theta}y + \dot{\theta}^2x
-\frac{\mu(\bar{r}+x)}{\left[(\bar{r}+x)^2+y^2+z^2\right]^{3/2}} + \frac{\mu}{\bar{r}^2}, \\
\ddot{y} &= -2\dot{\theta}\dot{x} - \ddot{\theta}x + \dot{\theta}^2y
-\frac{\mu y}{\left[(\bar{r}+x)^2+y^2+z^2\right]^{3/2}}, \\
\ddot{z} &= -\frac{\mu z}{\left[(\bar{r}+x)^2+y^2+z^2\right]^{3/2}}.
\end{align}

\subsubsection{Relative Attitude Kinematics and Rotational Dynamics with Unknown Inertia}
Let $\mathbf{\Gamma}(\mathbf{p})\in SO(3)$ map vectors from the target body frame to the chaser frame, parameterized by Modified Rodrigues Parameters (MRPs) $\mathbf{p}\in\mathbb{R}^3$. The relative angular velocity (expressed in the target frame) is
\begin{align}
\boldsymbol{\omega}_r &= \boldsymbol{\omega}_t - \mathbf{\Gamma}\boldsymbol{\omega}_c,
\end{align}
and the MRP kinematics are
\begin{equation}
\dot{\mathbf{p}}=\frac{1}{4}\Big[(1-\mathbf{p}^T\mathbf{p})\mathbf{I}_3
+2\mathbf{p}\mathbf{p}^T+2[\mathbf{p}\wedge]\Big]\boldsymbol{\omega}_r,
\end{equation}
where $[\mathbf{p}\wedge]$ is the standard cross-product matrix. The corresponding rotation matrix is computed via
\begin{align}
\epsilon_1 &= 4 \frac{1-\mathbf{p}^T\mathbf{p}}{(1+\mathbf{p}^T\mathbf{p})^2}, \quad
\epsilon_2 = 8 \frac{1}{(1+\mathbf{p}^T\mathbf{p})^2}, \\
\mathbf{\Gamma}(\mathbf{p}) &= \mathbf{I}_3 - \epsilon_1[\mathbf{p}\wedge] + \epsilon_2[\mathbf{p}\wedge]^2.
\end{align}

To enable inertia identification, the target inertia matrix is treated as unknown and estimated online. We parameterize the symmetric inertia tensor by its six independent elements
\begin{equation}
\boldsymbol{\theta}_J = \begin{bmatrix}J_{xx} & J_{yy} & J_{zz} & J_{xy} & J_{xz} & J_{yz}\end{bmatrix}^T,
\end{equation}
and define $\mathbf{J}_t(\boldsymbol{\theta}_J)$ accordingly. The relative Euler equation (torque-free with apparent/chaser terms as in \cite{phdthesis}) is written compactly as
\begin{equation}
\mathbf{J}_t \dot{\boldsymbol{\omega}}_r + \boldsymbol{\omega}_r \times (\mathbf{J}_t\boldsymbol{\omega}_r)
= \mathbf{M}_{app} - \mathbf{M}_{g} - \mathbf{M}_{ci}.
\end{equation}
In this work, $\boldsymbol{\theta}_J$ is modeled as a slowly-varying (random-walk) state, enabling gradual convergence without imposing an a priori calibrated inertia.

\subsection{Blender-Based ENVISAT Dataset and Sensor Simulation}
\label{subsec:blender}
A Blender-based simulation environment is developed to generate synchronized image and depth measurements under realistic orbital illumination. Unlike prior pipelines that relied on MATLAB-only camera models \cite{candan24}, the present workflow integrates (i) rigid-body propagation, (ii) camera projection, (iii) ray-cast visibility labeling, and (iv) photorealistic rendering within a single Blender/Python loop. The ENVISAT geometry is derived from ESA-provided CAD assets and simplified to emphasize the main bus edges and corners for robust keypoint tracking. A custom pinhole camera model is implemented using specified intrinsics $(f_x,f_y,c_x,c_y)$ and a fixed field of view. Each rendered frame stores: the 12-state relative truth (position/velocity, MRPs, angular rate), per-corner ground-truth pixel coordinates, per-corner depth, and a binary visibility flag computed via ray casting (line-of-sight test from the camera to each marker). Illumination is modeled using a directional light to approximate solar lighting, producing high-contrast shading and self-occlusion patterns. The simulation is rendered at 24~fps for Blender stability, while 1~Hz samples are used for Monte Carlo filtering experiments. Table~\ref{tab:camera_properties} summarizes the camera configuration used throughout the dataset generation.

\begin{table}[htbp] 
    \centering 
    \caption{Camera intrinsic properties and simulation parameters}
    \label{tab:camera_properties}
    \begin{tabular}{|l|l|} 
        \hline
        \textbf{Parameter} & \textbf{Value} \\
        \hline
        $FoV$ (Field of view) & 45$\degree$ \\
        \hline
        $f_x$ (Focal length in x-direction) & 1920 (pxs) \\
        \hline
        $f_y$ (Focal length in y-direction) & 1280 (pxs) \\
        \hline
        $c_x$ (Principal point x-coordinate) & 960 (pxs) \\
        \hline
        $c_y$ (Principal point y-coordinate) & 640 (pxs) \\
        \hline
        Affine translation (std. dev.) & 3 pixels (in both $x$ and $y$) \\
        \hline
        Affine rotation (std. dev.) & 1$\degree$ \\
        \hline
        Additive Gaussian noise (std. dev.) & 0.001 (normalized intensity) \\
        \hline
    \end{tabular}
\end{table}

\subsection{Corner Detection with CNN (TinyCornerNET)}
\label{subsec:cnn}
The perception front-end provides 2D corner measurements from monocular images. We adopt a lightweight heatmap-based keypoint detector, TinyCornerNET \cite{candanMdpi}, implemented as an encoder--decoder network with a ResNet-34 backbone \cite{unet, unetplus}. The network outputs a per-pixel likelihood map in which local maxima correspond to corner candidates. Training uses Blender-rendered images with projected ground-truth corner labels. A focal-style heatmap loss is employed to address the severe class imbalance between corner and background pixels, and only visibility-validated markers contribute to the loss to avoid penalizing occluded corners. Optimization is performed with AdamW optimizer \cite{adamw} and a fixed learning rate schedule.

\subsection{Augmented UKF for Joint Pose--Inertia Estimation}
\label{subsec:ukf}
The proposed estimator adopts the UKF to handle the strongly nonlinear relative translation--rotation coupling and the nonlinear camera projection. In contrast to the Extended Kalman Filter (EKF), which relies on local linearization of the dynamics and measurement models, the UKF approximates the propagation of mean and covariance by deterministically sampling a set of sigma points and passing them through the nonlinear functions. This preserves higher-order accuracy in the statistical mapping while retaining a linear update structure that remains practical for onboard implementation \cite{cavenagoDA}. The augmented UKF jointly estimates the relative kinematics and inertial parameters using the state,
\begin{equation}
\mathbf{x} = \begin{bmatrix}
\mathbf{r}^T & \mathbf{v}^T & \mathbf{p}^T & \boldsymbol{\omega}^T & \boldsymbol{\theta}_J^T & b
\end{bmatrix}^T,
\end{equation}
where $\boldsymbol{\theta}_J$ contains the six independent elements of the target inertia tensor and $b$ is a scalar depth-bias term.

\subsubsection{Sigma-Point Construction}
Let $L$ denote the dimension of the augmented state. The unscented transform is parameterized by $(\alpha,\beta,\kappa)$, with scaling
\begin{equation}
\lambda = \alpha^2(L+\kappa)-L.
\end{equation}
The corresponding weights for the sigma-point mean and covariance are
\begin{align}
W_m^{(0)} &= \frac{\lambda}{L+\lambda}, \\
W_c^{(0)} &= \frac{\lambda}{L+\lambda} + \left(1-\alpha^2+\beta\right), \\
W_m^{(i)} = W_c^{(i)} &= \frac{1}{2(L+\lambda)}, \quad i=1,\ldots,2L.
\end{align}
Given the prior estimate $(\hat{\mathbf{x}}_{k-1},\mathbf{P}_{k-1})$, sigma points are generated as
\begin{align}
\boldsymbol{\chi}_{k-1}^{(0)} &= \hat{\mathbf{x}}_{k-1},\\
\boldsymbol{\chi}_{k-1}^{(i)} &= \hat{\mathbf{x}}_{k-1} + \left[\sqrt{(L+\lambda)\mathbf{P}_{k-1}}\right]_i, \quad i=1,\ldots,L,\\
\boldsymbol{\chi}_{k-1}^{(i)} &= \hat{\mathbf{x}}_{k-1} - \left[\sqrt{(L+\lambda)\mathbf{P}_{k-1}}\right]_{i-L}, \quad i=L+1,\ldots,2L,
\end{align}
where $\sqrt{\cdot}$ denotes a matrix square root (e.g., Cholesky factorization).

\subsubsection{Time Update}
Each sigma point is propagated through the nonlinear state transition model,
\begin{equation}
\boldsymbol{\chi}_{k|k-1}^{(i)} = f\!\left(\boldsymbol{\chi}_{k-1}^{(i)},\mathbf{u}_{k-1}\right),
\end{equation}
where $f(\cdot)$ consists of the coupled relative translational and rotational dynamics, integrated using a 4th-order Runge--Kutta scheme. The predicted mean and covariance are recovered by weighted summation:
\begin{align}
\hat{\mathbf{x}}_{k|k-1} &= \sum_{i=0}^{2L} W_m^{(i)} \boldsymbol{\chi}_{k|k-1}^{(i)},\\
\mathbf{P}_{k|k-1} &= \sum_{i=0}^{2L} W_c^{(i)}
\left(\boldsymbol{\chi}_{k|k-1}^{(i)}-\hat{\mathbf{x}}_{k|k-1}\right)
\left(\boldsymbol{\chi}_{k|k-1}^{(i)}-\hat{\mathbf{x}}_{k|k-1}\right)^{T}
+\mathbf{Q},
\end{align}
where $\mathbf{Q}$ is the process-noise covariance. In particular, nonzero process noise is assigned to the inertia and bias states to avoid covariance collapse during constant-parameter estimation and to allow gradual convergence of $\boldsymbol{\theta}_J$ and $b$.

\subsubsection{Measurement Update}
For the update, propagated sigma points are mapped through the nonlinear RGB-D measurement function,
\begin{equation}
\boldsymbol{\gamma}_k^{(i)} = h\!\left(\boldsymbol{\chi}_{k|k-1}^{(i)}\right),
\end{equation}
where $h(\cdot)$ stacks per-marker measurements $\mathbf{z}_i=[u_i,\,d_i,\,v_i]^T$ generated by the corner geometry and camera projection, with depth modeled as $d_i = Y_i + b$ in the camera frame. The predicted measurement mean and innovation covariance are
\begin{align}
\hat{\mathbf{z}}_k &= \sum_{i=0}^{2L} W_m^{(i)} \boldsymbol{\gamma}_k^{(i)},\\
\mathbf{S}_k &= \sum_{i=0}^{2L} W_c^{(i)}
\left(\boldsymbol{\gamma}_k^{(i)}-\hat{\mathbf{z}}_k\right)
\left(\boldsymbol{\gamma}_k^{(i)}-\hat{\mathbf{z}}_k\right)^{T}
+\mathbf{R},
\end{align}
and the state--measurement cross-covariance is computed as
\begin{equation}
\mathbf{T}_k = \sum_{i=0}^{2L} W_c^{(i)}
\left(\boldsymbol{\chi}_{k|k-1}^{(i)}-\hat{\mathbf{x}}_{k|k-1}\right)
\left(\boldsymbol{\gamma}_k^{(i)}-\hat{\mathbf{z}}_k\right)^{T}.
\end{equation}
The Kalman gain and posterior update then follow:
\begin{align}
\mathbf{K}_k &= \mathbf{T}_k \mathbf{S}_k^{-1},\\
\hat{\mathbf{x}}_k &= \hat{\mathbf{x}}_{k|k-1} + \mathbf{K}_k\left(\mathbf{z}_k-\hat{\mathbf{z}}_k\right),\\
\mathbf{P}_k &= \mathbf{P}_{k|k-1} - \mathbf{K}_k \mathbf{S}_k \mathbf{K}_k^{T}.
\end{align}
Under the Gaussian assumption of the UKF posterior, the filter repeats the above time and measurement updates as new RGB-D observations become available for sensor fusion.

\subsection{Online Dual Adaptation of Measurement and Process Noises}
\label{subsec:adaptation}

\subsubsection{Innovation-Based Adaptive Measurement Covariance Tuning}
Measurement quality varies significantly with illumination, occlusion, and detector performance. To maintain consistency without manual retuning, we apply an innovation-based adaptive inflation in the measurement covariance. Let $\mathbf{e}_k=\mathbf{z}_k-\hat{\mathbf{z}}_k$ denote the innovation and $\mathbf{S}_k$ the predicted innovation covariance from the UKF. We compute
\begin{equation}
\mathbf{MTF}_k = \mathbf{e}_k\mathbf{e}_k^T - \mathbf{S}_k - \mathbf{R},
\end{equation}
retain only diagonal entries, assuming uncorrelated disturbances across channels, and enforce non-negativity element-wise:
\begin{equation}
\mathbf{MTF}_k \leftarrow \max(\mathbf{0}, \mathrm{diag}(\mathbf{MTF}_k)).
\end{equation}
The effective measurement covariance becomes $\mathbf{R}_{\mathrm{eff},k}=\mathbf{R}+\mathbf{MTF}_k$, which automatically down-weights unreliable measurements while preserving informative updates.

\subsubsection{Cross-Covariance-Guided Process Inflation}
During extended periods of degraded visibility or complete measurement loss, the filter must deliberately expand its process uncertainty to remain consistent; otherwise, the covariance can become overly optimistic, and the state may drift without sufficient correction. Rather than running a full UKF--RTS smoother, which would require a backward pass to refine past estimates, we exploit information already available from the forward UKF recursion, namely the cross-covariance between pre- and post-propagation sigma-point deviations. Specifically, using the sigma points at time $k\!-\!1$ and their propagated counterparts at time $k$, we construct the forward cross-covariance
\begin{equation}
\mathbf{D}_{k}
= \sum_{i=0}^{2L} W_c^{(i)}
\left(\boldsymbol{\chi}_{k-1}^{(i)}-\hat{\mathbf{x}}_{k-1}\right)
\left(\boldsymbol{\chi}_{k|k-1}^{(i)}-\hat{\mathbf{x}}_{k|k-1}\right)^{T},
\end{equation}
where $W_c^{(i)}$ are the unscented transform covariance weights. Intuitively, $\mathbf{D}_k$ captures how perturbations in the state at $k\!-\!1$ map into perturbations after propagation to $k$, and therefore encodes the dominant directions of uncertainty growth induced by the dynamics. When no measurement update is performed (e.g., insufficient visible markers), we adapt the process uncertainty online through an additive covariance injection
\begin{align}
\mathbf{Q}_{\mathrm{adaptive},k} &= \mathbf{D}_k\left(\mathbf{P}_{k|k-1}-\mathbf{P}_{k-1}\right)\mathbf{D}_k^{T}, \\
\mathbf{P}_{k} &= \mathbf{P}_{k|k-1} + \mathbf{Q}_{\mathrm{adaptive},k}.
\end{align}
This update is motivated by smoothing intuition \cite{Sarkka2013}: the increment $\mathbf{P}_{k|k-1}-\mathbf{P}_{k-1}$ measures how much uncertainty has grown during propagation, while $\mathbf{D}_k$ acts as a projection that aligns this growth with the directions along which the sigma points actually diverged. Consequently, the injected covariance is directionally informed, increasing uncertainty primarily along the most sensitive modes associated with model mismatch and unobserved dynamics. The resulting procedure is fully online, non-iterative, and requires no batch statistics or hand-tuned outage windows, making it well-suited for onboard implementation and particularly effective under eclipse-like dropouts or severe occlusion events.

\section{Results}
\begin{figure}[H]
    \centering
    \includegraphics[width=\linewidth]{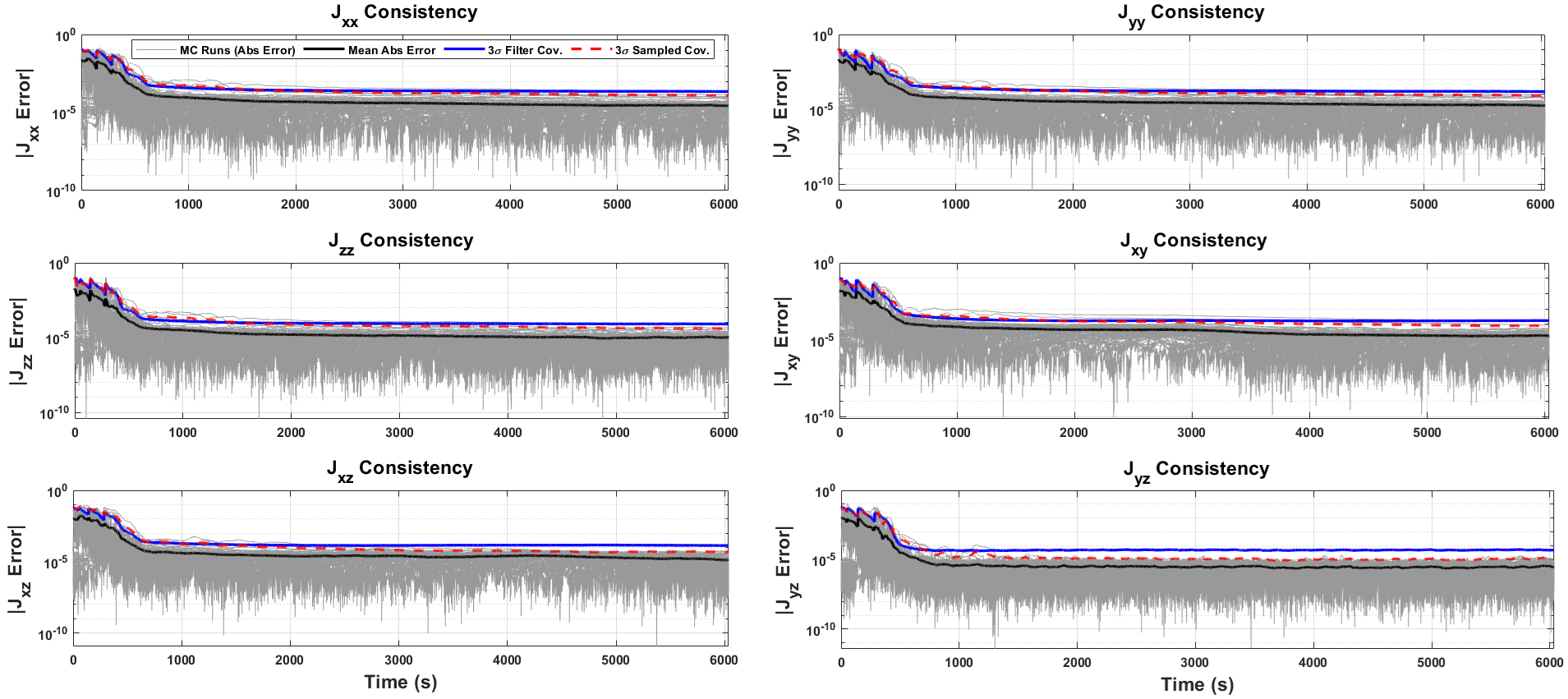}
    \caption{Monte Carlo convergence of the trace-normalized inertia states ($N=100$)}
    \label{fig:1}
\end{figure}

Two initialization settings were considered in the Monte Carlo study to assess both nominal convergence and robustness to prior uncertainty. In the first setting, the inertia state was initialized using an informed but uncertain prior. The initial normalized diagonal inertia components were centered at the nominal target values, $\bar{J}_{xx}\approx 0.063$, $\bar{J}_{yy}\approx 0.459$, and $\bar{J}_{zz}\approx 0.478$, while the cross-product terms were assigned zero mean. For each Monte Carlo run, the initial inertia estimate was generated by applying Gaussian perturbations to this prior according to the assumed initial covariance. The sampling employed the full prescribed one-standard-deviation prior uncertainty, before trace normalization, to standard deviations of $0.06$ for $\bar{J}_{xx}$, $\bar{J}_{yy}$, and $\bar{J}_{zz}$, $0.05$ for $\bar{J}_{xy}$, and $0.03$ for $\bar{J}_{xz}$ and $\bar{J}_{yz}$. This initialization represents a realistic scenario in which approximate prior knowledge is available, but the initial inertia estimate remains subject to substantial uncertainty. As shown in Figure \ref{fig:1}, the filter demonstrates consistency and rapid convergence. The steady-state standard deviation across the Monte Carlo trials remained remarkably low, specifically below $0.07\%$ for the primary diagonal components ($\bar{J}_{xx}$, $\bar{J}_{yy}$, $\bar{J}_{zz}$). While the cross-product terms exhibited slightly higher relative variance (e.g., $1.8\%$ for $\bar{J}_{xy}$), the absolute mean bias for all terms remained on the order of $10^{-5}$ to $10^{-7}$ in normalized units. This highlights the estimator's ability to precisely identify the target's inertial properties despite the initial uncertainty.

\begin{figure}[H]
    \centering
    \includegraphics[width=\linewidth]{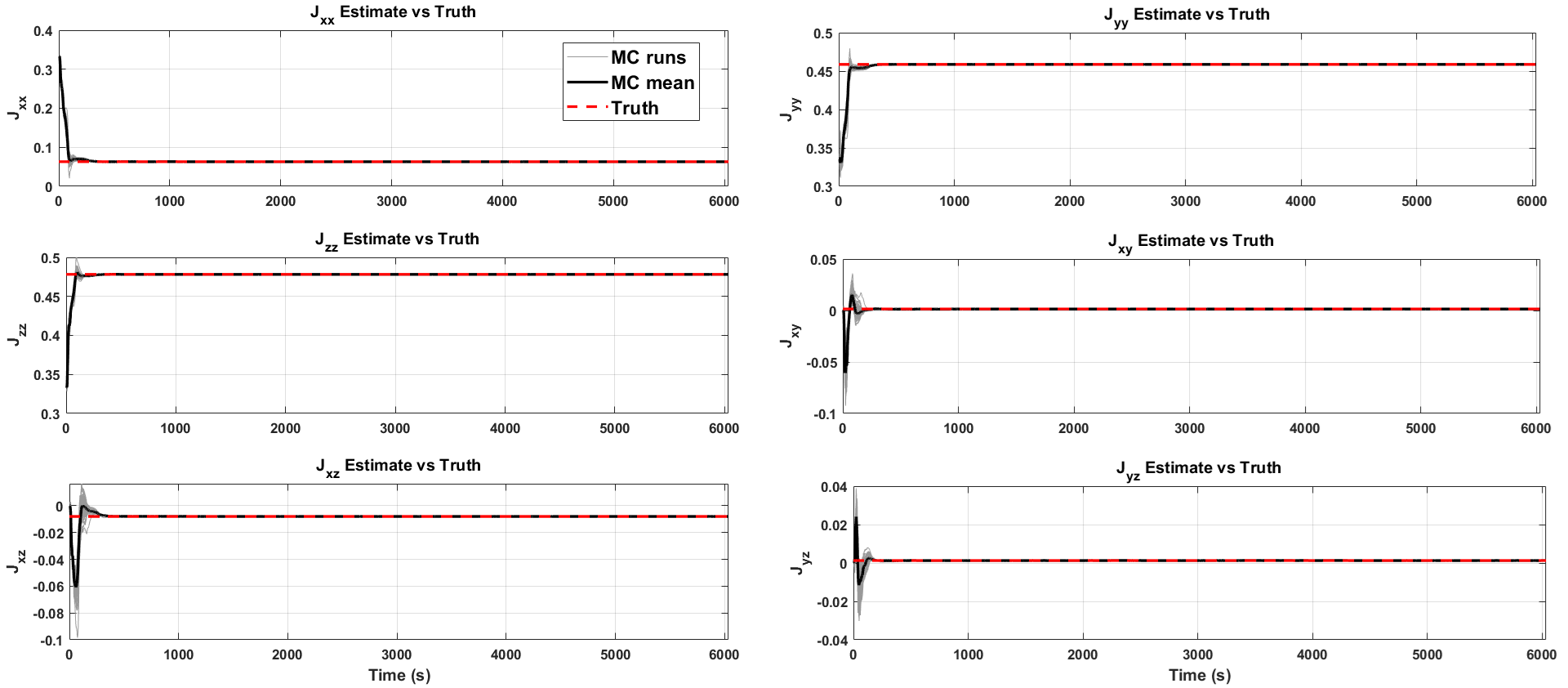}
    \caption{Monte Carlo convergence ($N=100$) under uninformative isotropic initialization}
    \label{fig:2}
\end{figure}

To further evaluate robustness, a second set of Monte Carlo trials used a fully uninformative initialization. In this case, the target inertia tensor was treated as completely unknown at the initial time, and the filter was initialized with the isotropic normalized inertia guess
$\bar{J}_{xx}=\bar{J}_{yy}=\bar{J}_{zz}=1/3$ and
$\bar{J}_{xy}=\bar{J}_{xz}=\bar{J}_{yz}=0$.
This corresponds to an inertia distribution with no preferred principal axis and no prior knowledge of cross-products. By avoiding any bias toward the true inertia, this case provides a stricter test of the proposed estimator and highlights its ability to recover inertial parameters online using only RGB-D measurements and rigid-body dynamics. As illustrated in Figure \ref{fig:2}, all 100 runs converge rapidly to the true values within the first 500 seconds, despite the initial isotropic assumption. The steady-state performance remains exceptional; the standard deviation across all trials is below $0.05\%$ for the primary diagonal components. The mean bias remains negligible, on the order of $10^{-5}$ to $10^{-8}$ normalized units, confirming that the estimator is asymptotically unbiased even when starting from a worst-case prior.

\section{Conclusion}
This paper presented an augmented UKF framework for the simultaneous estimation of the 6-DOF relative pose and the full inertia tensor of a non-cooperative target spacecraft. While the augmented UKF jointly estimates the full 12-state relative kinematics, this study focuses exclusively on the recovery of the inertial parameters. The performance of the 6-DOF pose estimation component, including its convergence and accuracy under various illumination conditions, has been extensively documented in our previous work \cite{candanMdpi}. The primary novelty of the current study lies in the online identification of the full inertia tensor from those kinematic observations; therefore, the pose estimation results are omitted in this work for brevity. By fusing monocular vision-based keypoint detections with LiDAR-assisted depth information, the proposed estimator dynamically recovers the target’s mass distribution online without requiring a priori knowledge of the mass properties. The integration of adaptive measurement inflation and cross-covariance-guided process noise ensures numerical stability and consistency. Numerical validation via high-fidelity Blender simulations and extensive Monte Carlo trials demonstrated the estimator's precision and robustness. Even when starting from a fully uninformative isotropic prior, the filter successfully identified the target's inertial parameters within 500 seconds, achieving a steady-state standard deviation of less than $0.05\%$ for the primary diagonal components.

These findings suggest that the proposed architecture is a viable solution for the long-horizon relative navigation required in active debris removal and on-orbit servicing missions. By accurately recovering the target's physical parameters online, the framework enables reliable trajectory prediction and robust guidance under high parametric uncertainty. Future work will focus on expanding the state vector to include mass estimation under known thruster excitation and validating the pipeline against experimental laboratory data involving representative satellite mockups.

\bibliography{sample}

\end{document}